\pdfoutput=1

\documentclass[11pt]{article}

\usepackage{naacl2021}

\usepackage{times}
\usepackage{latexsym}
\usepackage{graphicx}

\usepackage[T1]{fontenc}

\usepackage[utf8]{inputenc}

\usepackage{microtype}

    \title{Natural Language Processing 4 All (NLP4All): A New Online Platform for Teaching and Learning NLP Concepts}

\author{Rebekah Brita Baglini \\
  Aarhus University \\
  Interacting Minds Center \\
  \texttt{rbkh@cc.au.dk} \\\And
  Hermes Arthur Hjorth \\
  Aarhus University \\
  ScienceAtHome \\
  \texttt{arthur@mgmt.au.dk} \\}

\begin{document}
\maketitle
\begin{abstract}
Natural Language Processing offers new insights into language data across almost all disciplines and domains, and allows us to corroborate and/or challenge existing knowledge. The primary hurdles to widening participation in and use of these new research tools are, first, a lack of coding skills in students across K-16, and in the population at large, and second, a lack of knowledge of how NLP-methods can be used to answer questions of disciplinary interest outside of linguistics and/or computer science.  To broaden participation in NLP and improve NLP-literacy, we introduced a new tool web-based tool called Natural Language Processing 4 All (NLP4All). The intended purpose of NLP4All is to help teachers facilitate learning with and about NLP, by providing easy-to-use interfaces to NLP-methods, data, and analyses, making it possible for non- and novice-programmers to learn NLP concepts interactively. 

\end{abstract}

\section{Introduction}

An emerging body of work has explored ways of lowering the threshold for people to work with AI and ML-technologies, specifically in educational contexts. Much of this work has focused on making AI “explainable” \cite{gunning_explainable_2017} by visualizing the underlying math, or visualizing how machines make decisions. However, NLP has been largely absent from these efforts so far. To address this gap, we developed a new educational tool called NaturalLanguageProcessing4All (NLP4All), which allows teachers to interactively introduce applications of statistical NLP to students without any coding skills.


NLP4All\footnote{A demo version of NLP4All can be accessed here: http://86.52.121.12:5000/, pre-loaded with the data and analysis described in this paper.} is a web-based interface for teaching and learning NLP concepts, designed with flexibility and accessibility in mind. It is an open source application built in Python on top of the Flask framework, and can therefore be easily extended with existing Python-based NLP- and ML-packages. The first prototype of NLP4All is designed to work with tweets only, but we are currently expanding to be able to work with any kind of text, bringing NLP tools to a wider array of disciplines and student populations. 

We first present the design of the tool and its current capabilities, and then briefly describe two different real-world settings in which we have used NLP4All.

\section{Design}

NLP4All is an interactive tool which is designed to supplement classroom modules introducing NLP or machine learning concepts for non-specialists. The platform facilitates hands-on experimentation with NLP applications using real data, without requiring students to do any programming. 

NLP4All provides two different user types: teachers and students. Whereas teachers can see data from all students and can create new projects, students’ activities are more limited in the system.

The system is organized into \textbf{user groups}, \textbf{projects}, and \textbf{analyses}. To better describe how the system, we will briefly outline how each of these work.

\subsection{User Groups}
User groups provide an easy way to organize groups of students. A group will consist of students who are doing the same activities, and simply act as an easy way to add students to projects. They will typically consist of students in one class. User groups can be created by a teacher and associated with a unique sign-up link to be distributed to the intended recipient group.

\subsection{Projects}
Projects in NLP4All offer teachers a way to organize a lesson by selecting some texts of interest, and tying them to a user group. A project consists of a title, a description, a user group (of students), and a set corpora that will be included in the project. Teachers create projects with associated datasets prior to classroom sessions; they can either choose to use several pre-loaded datasets (Tweets from the accounts of different American and Danish politicians or political parties)  or upload their own texts in .csv or .json format. 

\begin{figure}[h!]
    \centering
    \includegraphics[width=8cm]{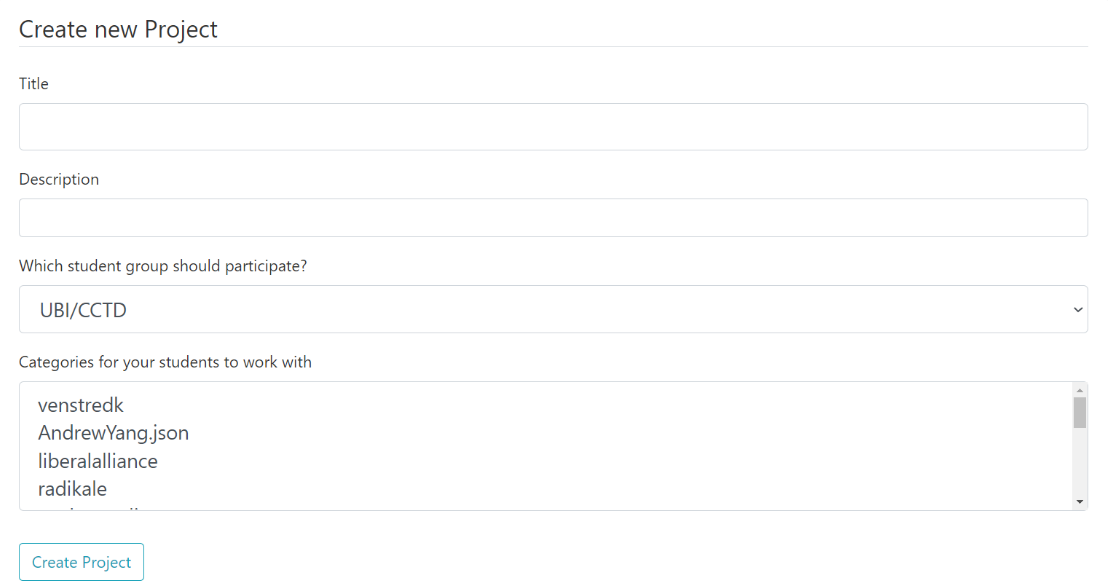}
    \caption{Teacher view displaying Projects interface.}
    \label{fig:projects}
\end{figure}

\subsection{Analyses} 

Inside a project, both teachers and students can create a new analysis. An analysis is NLP4All’s name for an untrained model and all data associated with training it. Students can create a new analysis if they think that they have trained their old model poorly and want to start from scratch. Students can only create personal analyses - i.e. analyses that are unique to their account, and not shared among other members of the user group. Teachers, in contrast, can create analyses that are shared between all members of a user group. 

\begin{figure}[h!]
    \centering
    \includegraphics[width=6cm]{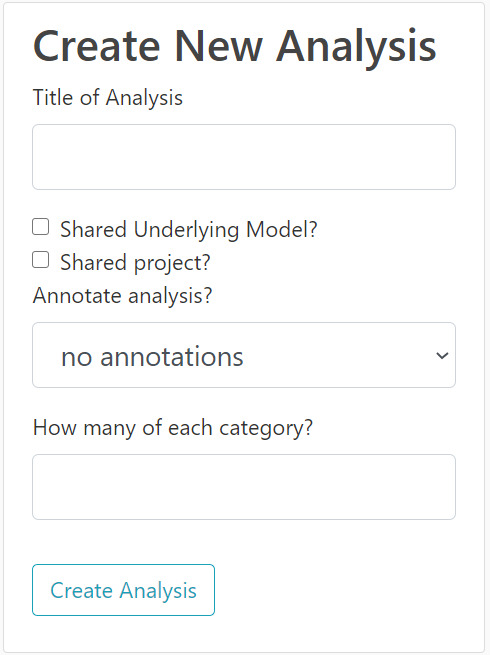}
    \caption{New Analysis interface.}
    \label{fig:analysis}
\end{figure}

There are two different ways in which an analysis can be shared (Fig. \ref{fig:analysis}): the teacher can choose to just share the texts that students hand-label, or they can choose to share an underlying  model. For the former case, the teacher can specify a number of texts from each category, and NLP4All will create a mini-corpus of just those texts for students to work with. For the latter case, students work with the whole corpus of the texts present in the project, but all train the same underlying model as they hand label texts.

NLP4All also supports text annotation, but we do not discuss this functionality here.

\section{Example: Teaching text classification with NLP4All} 

The initial version of NLP4All features interactive tools for a curriculum module on \textbf{text classification} with Naïve Bayes and Logistic Regression models.
In this section, we walk through an example of how NLP4All could be used in the classroom to introduce Naïve Bayes text classification using a corpus of posts from the Twitter accounts of Joe Biden and Bernie Sanders collected in the run-up to the 2020 Democratic Primaries. 

\begin{figure}[h!]
    \centering
    \includegraphics[width=8cm]{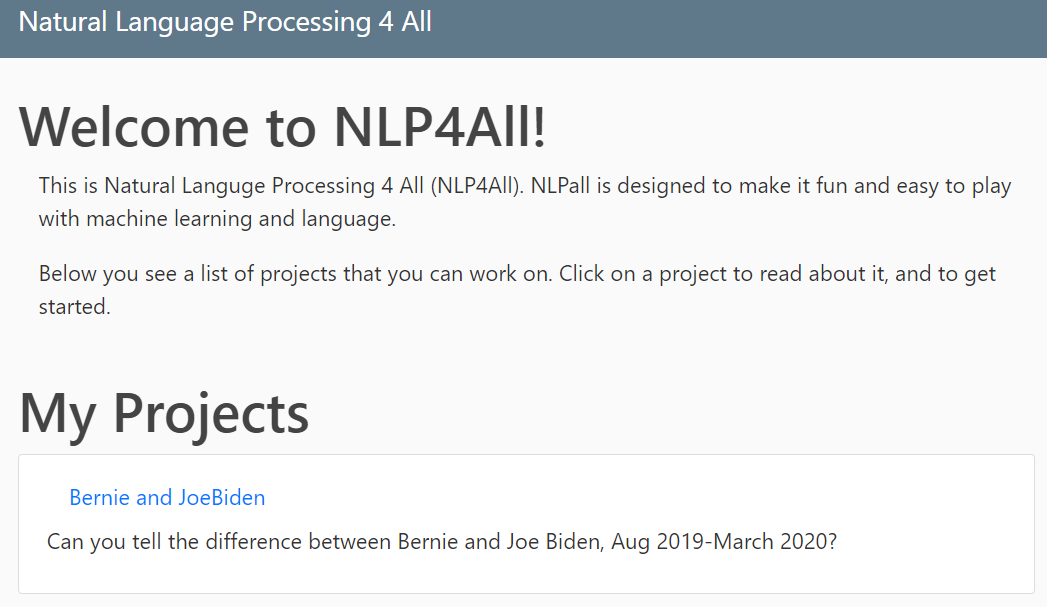}
    \caption{Landing page for students.}
    \label{fig:landing}
\end{figure}

Upon logging in, students see a landing page which lists all projects that the student is currently part of, as determined by the instructor (Fig. \ref{fig:landing}).  

\subsection{Hand labeling: The Tweet View}

\begin{figure}[h!]
    \centering
    \includegraphics[width=8cm]{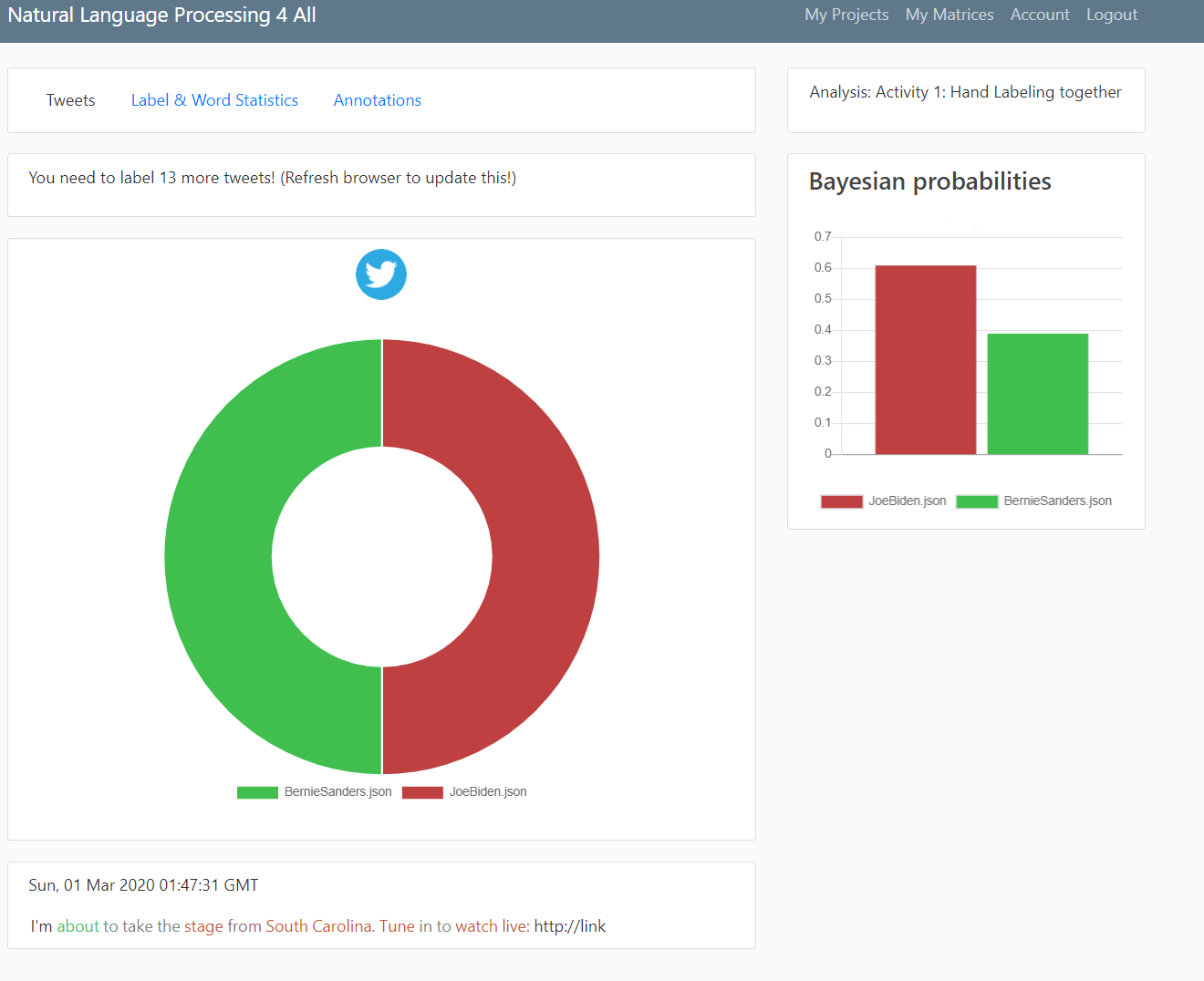}
    \caption{The Tweet View interface. }
    \label{fig:tweet}
\end{figure}

The current implementation of NLP4All has a special view called the Tweet View, where students hand label Twitter data, as seen in Fig. \ref{fig:tweet}. At the bottom of the page, it shows the tweet currently being labeled.  Students label each tweet by dragging the Twitter bird to whichever side of the circle represents the category that they think the it belongs to. For example, by dragging the bird to the green part of the circle, a student would label the tweet in Fig. \ref{fig:tweet} as having been written by Bernie Sanders. All Tweets in this dataset were pre-processed so that mentions, hashtags, and links were replaced with \@mention, \#hashtag, and http://link, respectively.

In the top right corner Fig. \ref{fig:tweet}, the Tweet View also shows the model’s best estimate of who wrote the tweet, based on the data that it has been trained on so far. For the current tweet, the model estimates that Biden wrote the tweet with around .63 and that Bernie wrote it with the remaining .37.  By making the classification explicit to students, we hope to achieve two different goals. First, we want students to understand how each word contributes to the overall classification. Second, we want students to critically reflect on whether they agree with the model's assessment. It is, of course, important when working with ML to not always trust your model. By giving students a clear idea of how the model reaches its conclusions, we hope that students can learn not only to be skeptical of ML models in a general sense, but that they can begin to understand why particular features or combinations of features may confuse a model, and through this get a better sense of what can go wrong when ML makes classifications. Finally, at the top of the page, students are shown how many more tweets this student needs to hand label before they are done with all tweets.

\section{Using label and word statistics to facilitate learning} 

Clicking the \emph{Label \& Word Statistics} link at the top of the page gives an overview of all tweets that students in this shared analysis have labeled, how many labeled them correctly, and how many labeled them incorrectly.

The purpose of this view is for the teacher to be able to discuss with students which tweets are hard to classify (i.e. the ones that many hand label incorrectly), which ones are easy, to foster discussion in the classroom what we know about these data. The screenshot below shows us this list, sorted by correct-\% in ascending order. Here, we see that 22 out of 23 students mislabeled the tweet with the text, “FLORIDA: Today is the LAST DAY to register for the Democratic primary. You must register as a Democrat to vote in the March 17th primary at http://link Let's win this together! http://link”. This is not surprising, given how generic this tweet is given the choice between two democratic candidates competing in the same primaries.

\begin{figure}[h!]
    \centering
    \includegraphics[width=8cm]{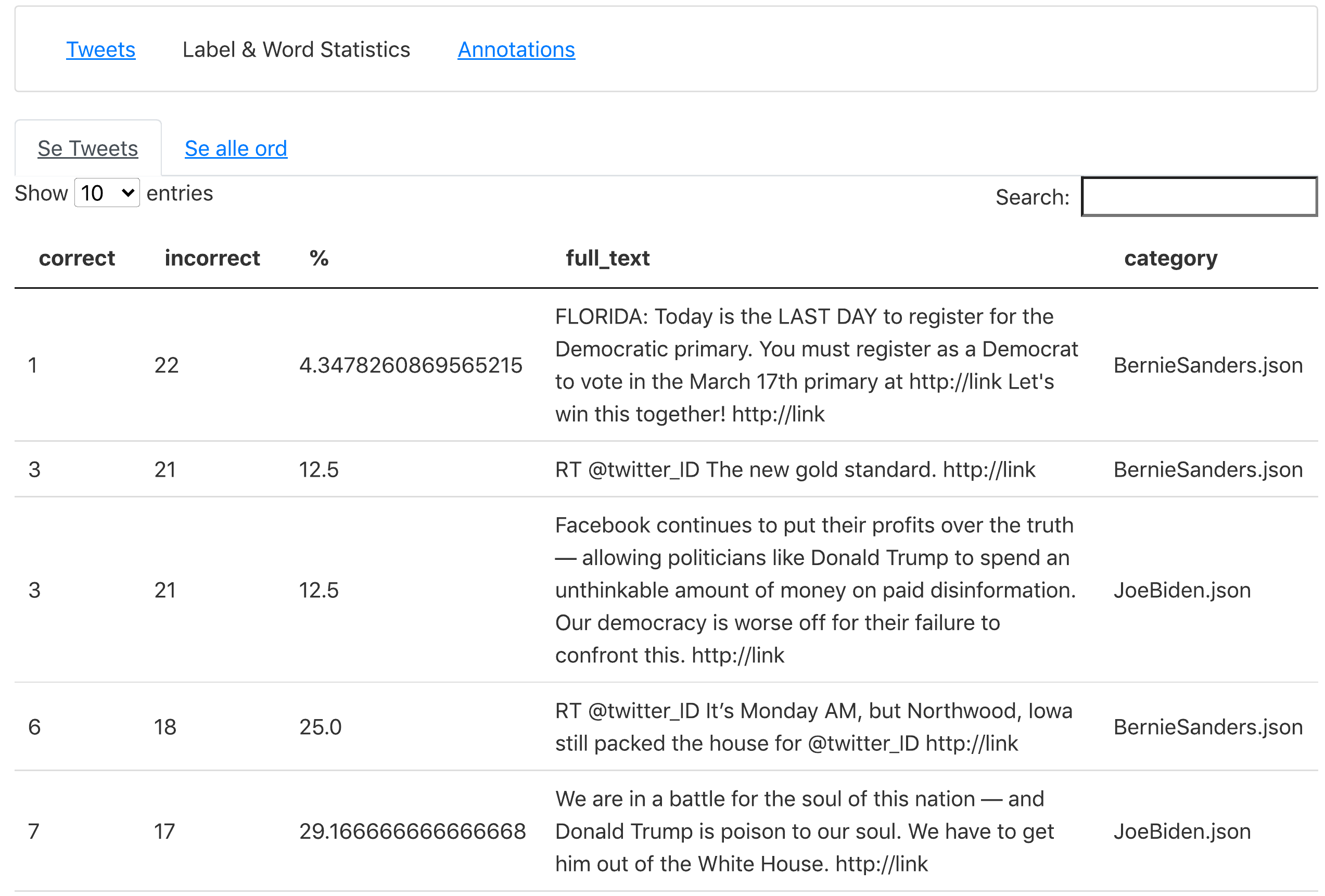}
    \caption{Label \& Word Statistics view. After students participate in hand labeling data, this view can be used to facilitate discussion.}
    \label{fig:labelwordstats}
\end{figure}

Sorting the list in descending order of correct-\%, we can see the tweets that all or most students labeled correctly. For instance, the tweet “We will not defeat Donald Trump with a candidate who, instead of holding the crooks on Wall Street accountable, blamed the end of racist policies such as redlining for the financial crisis.” was seemingly easily recognizable as a Bernie-tweet (23/23 students labeled it correctly.) Similarly, “We need a leader who will be ready on day one to pick up the pieces of Donald Trump's broken foreign policy and repair the damage he has caused around the world. http://link” was easily recognizable by 23 students as having been written by the Biden team. (The wrong guess was from the teacher demonstrating the system to students.) 

\begin{figure}[h!]
    \centering
    \includegraphics[width=8cm]{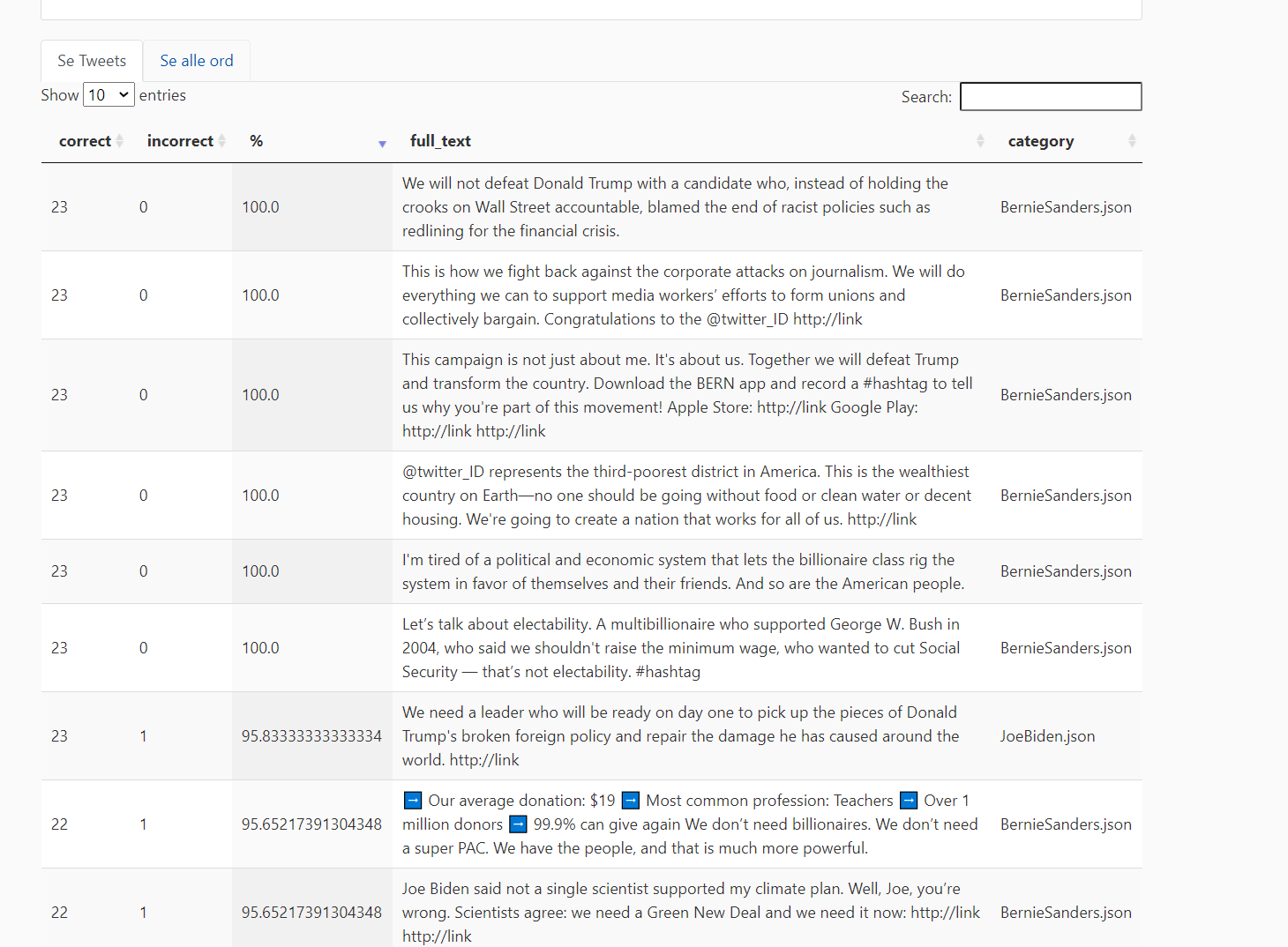}
    \caption{Label \& Word Statistics view, sorted by \% correctly labeled.}
    \label{fig:labelwordstatssorted}
\end{figure}

The \emph{See all words} link  brings up a table of all words present in tweets that have been labeled so far. This list shows how many times each word has appeared in the set of labeled tweets, and to which extent each word predicts each of the categories in the project, here Joe Biden and Bernie Sanders. For instructional purposes, this list can be used to discuss a variety of questions. For example, in Fig.  \ref{fig:labelwordstatssorted}, the list is sorted by how many times a word appears. Since the texts have not been filtered, this can act as a point of entry to a  discussion of why only some words are meaningful when it comes to distinguishing between different classes. The teacher may choose this moment to introduce students to the concept of stop words, or even to the notion of statistical power laws behind word frequency (Zipf's law).

\begin{figure}[h!]
    \centering
    \includegraphics[width=8cm]{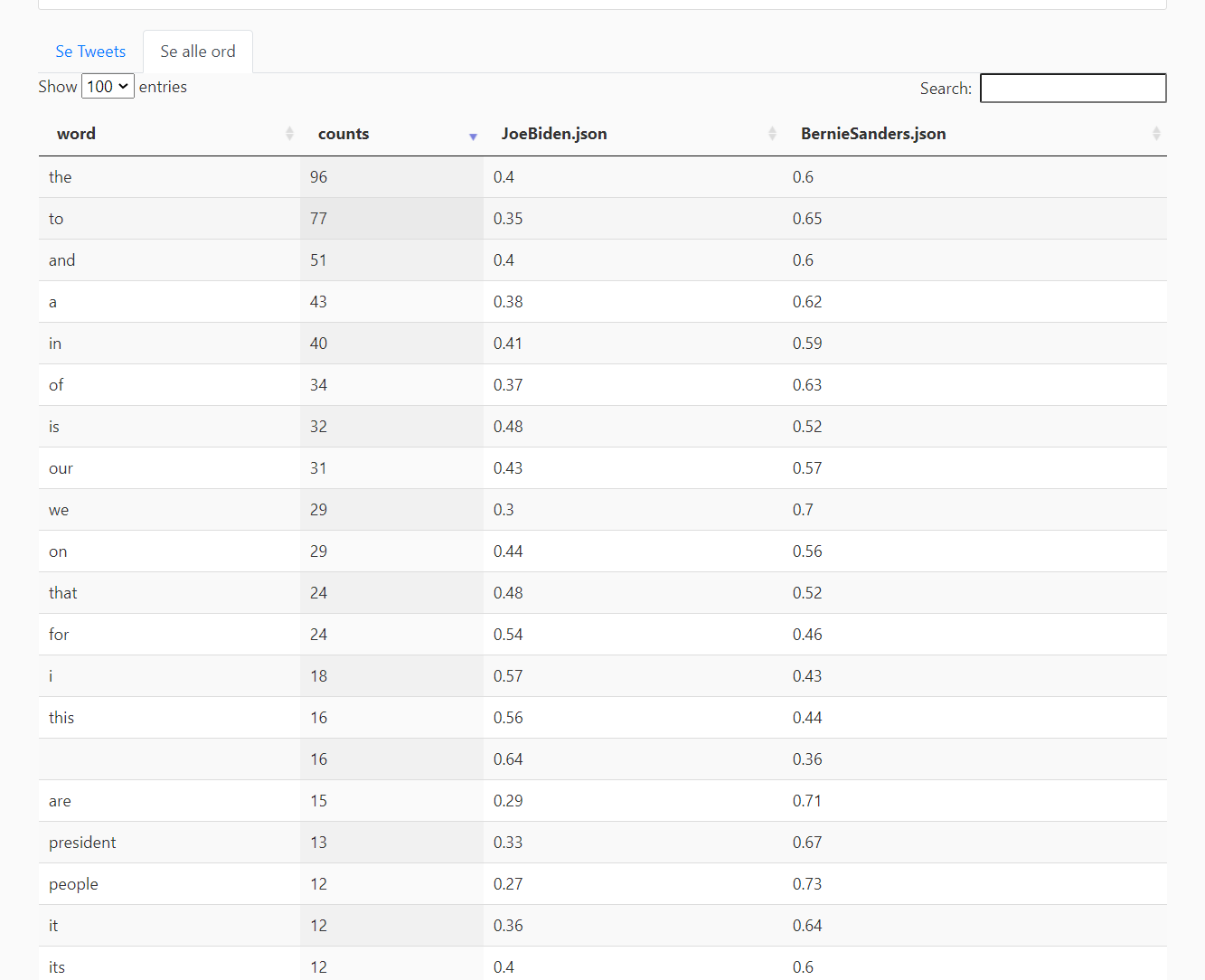}
    \caption{Word statistics, sorted by frequency.}
    \label{fig:wordstats}
\end{figure}

\subsection{Model training and evaluation} 

\begin{figure}[h!]
    \centering
    \includegraphics[width=8cm]{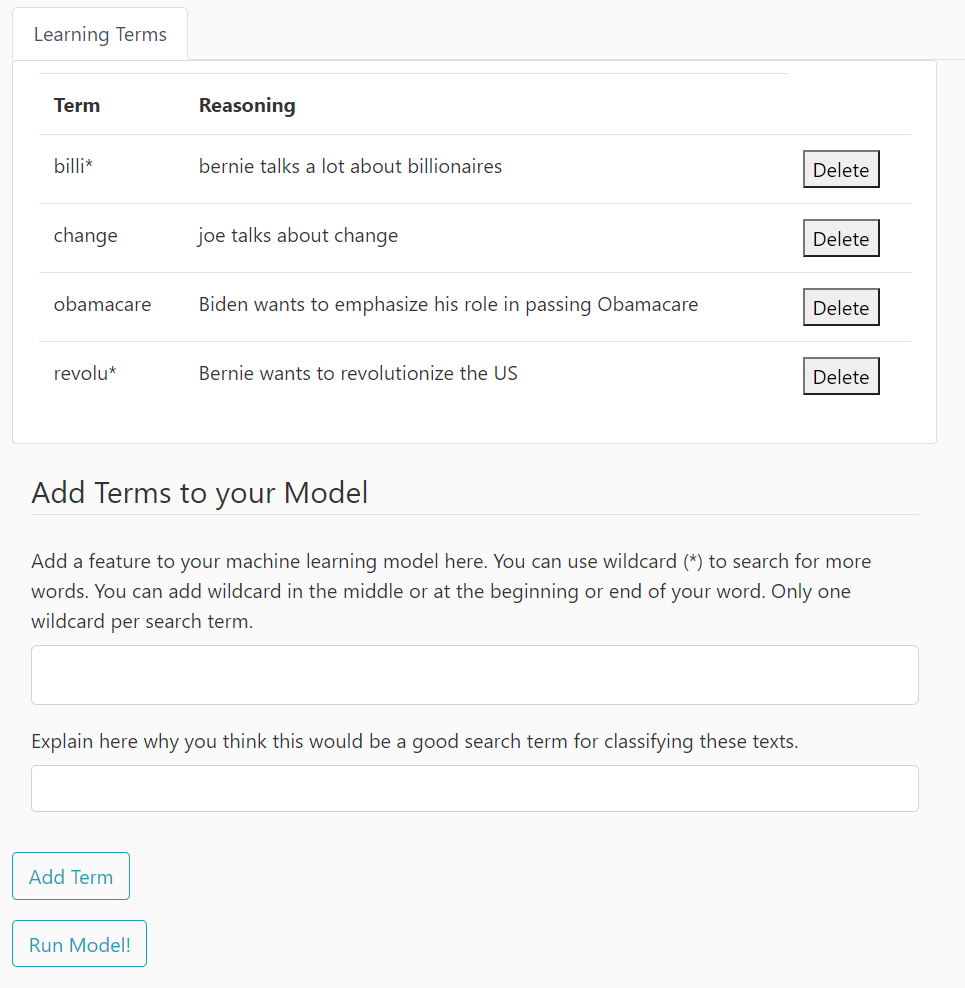}
    \caption{Student-defined word features}
    \label{fig:terms}
\end{figure}

NLP4All also lets students create their own Naïve Bayes models by specifying a set search terms to train their model on. Asterisks work as wildcards, and can be placed anywhere in a word, including at the front or back.  Importantly, for the purpose of reflection and classroom discussions, students are prompted to also state a reason why they think this would be a good word feature for distinguishing between the categories of text in the project. In the screenshot in Fig. \ref{fig:terms} we see how one student has added four different terms and their reasons for inclusion.

\begin{figure}[h!]
    \centering
    \includegraphics[width=8cm]{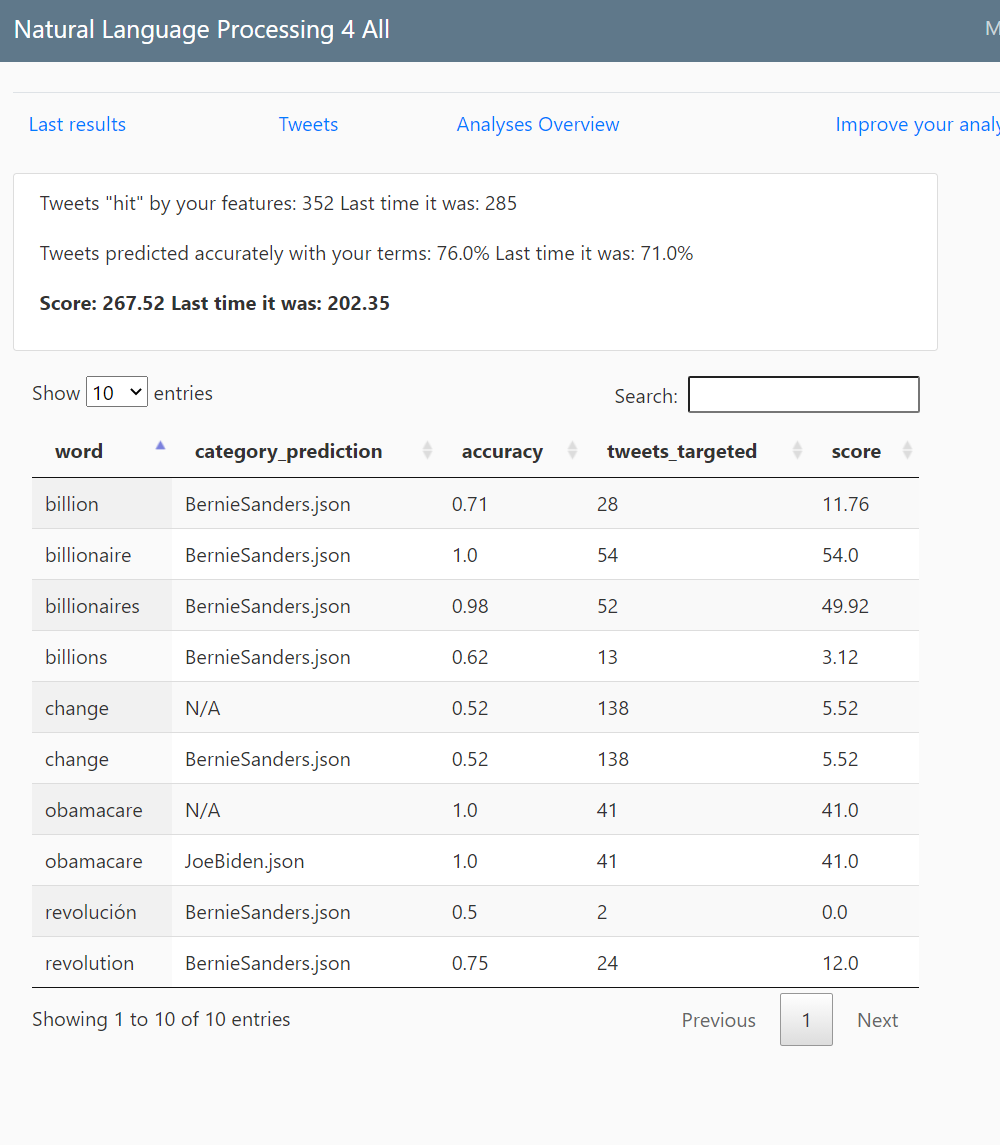}
    \caption{Feedback on model performance}
    \label{fig:feedback}
\end{figure}

By clicking \emph{Run Model} button at the bottom, NLP4All finds all words that match the search terms (including wild cards) and trains a Naïve Bayes model based on them. It returns the screen shown in the example screenshot in Fig. \ref{fig:feedback}. 

Here, the user is shown a table with information on each word found from their set of search terms: which category the model predicts based on the training set, how accurately that word was for predicting tweets in the test set, and how many tweets contain the word (‘targeted’) in the test set. 

In this particular case, we see that the word `billionaire' is the most predictive term: based on the training set, the model predicts that a tweet containing `billionaire' is written by Bernie Sanders, and this was the case in every single one of the 54 tweets containing this word in the test set.  Finally, each search term earns a score. This provides a “gamified” element of NLP4All that can be ignored, but that has been found to be motivating and fun to many students. Students can iterative improve their models by adding or removing search terms, and running these analyses until they are happy with the terms they have found.

\begin{figure}[h!]
    \centering
    \includegraphics[width=8cm]{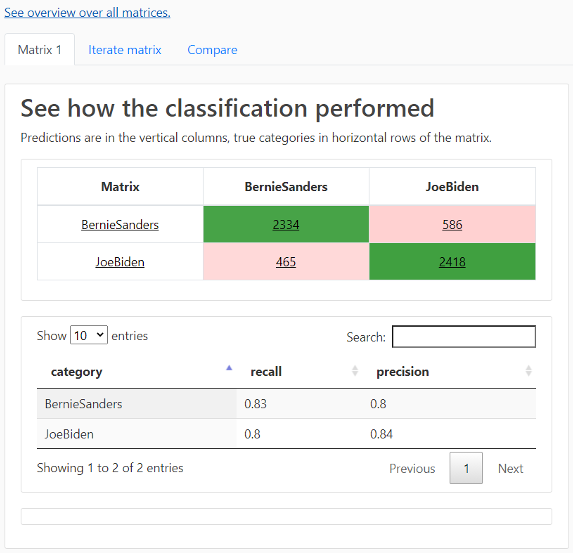}
    \caption{Viewing a confusion matrix}
    \label{fig:matrix}
\end{figure}

NLP4All users can also view confusion matrices for any of their trained models, as illustrated in Fig. \ref{fig:matrix}.

\section{NLP4All in the classroom: Case studies}


\subsection{Introducing text classification to MA students in the Humanities} 

Recent revisions to the national study regulations for humanities students in Denmark place an emphasis on digitization and digital literacy. As a result, study programs in Humanities and Social Sciences are adding courses like Text Analytics, Computational Linguistics, and Data Science to their standard curriculum. This poses a challenge, however, as students in these programs typically have little background or interest in math, statistics, or programming and some lack even basic computer skills. 

Working with a faculty member in at a major Danish university, we developed a classroom module on Classification as part of a introductory course on Computational Linguistics. The students in the class were second semester masters students in Linguistics and Cognitive Semiotics. Only one out of 22 students had any background in programming, and none had taken a specialization in math or science in high school.  

Student prepared for the two-week module by reading an introductory textbook chapter on Document Classification \cite{dickinson_language_2012} covering the Naive Bayes algorithm and completing the exercises at the end of the chapter. These conceptual and mathematical foundations were reinforced and built upon in two one-hour classroom lectures, which also introduced the contrast between generative and discriminative models and classification with logistic regression. In the second two hours of each three-hour class meeting, the class collectively participated in training and evaluating classification models with NLP4All using the Biden/Bernie Twitter data described above. The activities were broken into time blocks with discussion following each stage. 

In a post-classroom evaluation of students (\emph{n}=20), 100\% agreed that `the in-class exercises using NLP4All were effective for learning' and that the exercises `improved my understanding of text classification'. In additional comments, several students reported that they enjoyed the gamified and competitive aspect of NLP4All, while others mentioned that they liked the opportunity to work with real-world social media data. 

\subsection{Facilitating social studies discussion in a Danish high school}  

We tested NLP4All in a Social Studies high school classroom. In collaboration with a social studies teacher, we developed a 6-hour learning unit on language, ideology and political parties. The unit was designed to address one of learning goals of our national learning standards, specifying that students should learn about the different policy positions of political parties (we have 13 in our national parliament.) In other words, the purpose was not to teach NLP, but to teach with NLP, and to offer NLP-methods as a way of analyzing larger amounts of text than is otherwise possible.  

24 2nd year (sophomore) Danish high school students participated, with roughly equal numbers of girls and boys. In a survey sent out to students prior to our test, none of these students self-reported as having any programming experience, and 20 out of 24 reported no or low interest in computer science or machine learning. All had self-selected into “A-level” Social Studies, a 3-year elective class. About one third of students had immigrant backgrounds, slightly above the national average. 

The teacher and students used NLP4All to discuss tweets from pairs of Danish political parties. First, students had to label tweets and a model to tell a socialist and a nationalist party apart. Then, students did the same with the same socialist party, and a libertarian party. 

We cannot report on more concrete findings or analyses of learning data at present, as these results are currently under review at another venue. However, in evaluations the students reported enjoying being able to provide concrete evidence for their analyses. To them, purely qualitative analyses sometimes feel fluffy, but by showing that their analyses were backed up by hundreds or thousands of tweets made them feel more comfortable making claims during classroom discussions. 

\section{Comparison to Prior Work}
We have found two systems that are similar to NLP4All in certain ways, though also different in others. GATE \cite{cunningham_gate_2002} is a combined Java API  and graphical interface that makes it easy to create NLP pipelines without writing all code from scratch. While it was originally made for researchers, it has also  been used in teaching contexts \cite{bontcheva_using_2002} because it makes it easy for novice programmers to implement more sophisticated NLP methods than they could do on their own. However, presumably because GATE was made for researchers, it is not made for classroom contexts, and does not offer interfaces on data that would be useful for teachers during the teaching situation. \citet{light2005web} present a web interface that lets novices process text with common models and methods like NLTK’s PoS tagger and grammar parser. The web interface lets novices combine these models when processing text and visualizes output. However, similar to GATE, this interface does not provide views on data that are relevant to the teaching context. Additionally, the modules are black boxed to the students and do not provide any information on how the models work, how they are trained, or how they make predictions. 

\section{Conclusion}

At present, NLP4All provides support for teaching the following technical topics, without requiring any programming on the part of teachers or students: 

\begin{itemize} 
\item Classification algorithms
\begin{itemize} 
\item Naive Bayes 
\item Logistic regression
\end{itemize} 
\item Feature selection 
\item Supervised machine learning, test and train sets
\item Model evaluation 
\begin{itemize}
    \item Precision, recall, f-measure
    \item Confusion matrices
\end{itemize}
\end{itemize}
With a grant received in Spring 2021, the platform will be extended to support new learning modules on tf-idf, vector-based representations of texts, topic modeling, and word embeddings. 

\section*{Acknowledgments}
This work received Seed Funding from the Interacting Minds Center and additional support from the Department of Linguistics, Cognitive Science, and Semiotics at Aarhus University. 

\bibliography{NLP4all,custom}

\begin{thebibliography}{5}
\expandafter\ifx\csname natexlab\endcsname\relax\def\natexlab#1{#1}\fi

\bibitem[{Bontcheva et~al.(2002)Bontcheva, Cunningham, Tablan, Maynard, and
  Hamza}]{bontcheva_using_2002}
Kalina Bontcheva, Hamish Cunningham, Valentin Tablan, Diana Maynard, and Oana
  Hamza. 2002.
\newblock Using {GATE} as an {Environment} for {Teaching} {NLP}.
\newblock In \emph{Proceedings of the {ACL}-02 {Workshop} on {Effective} tools
  and methodologies for teaching natural language processing and computational
  linguistics}, pages 54--62.

\bibitem[{Cunningham(2002)}]{cunningham_gate_2002}
Hamish Cunningham. 2002.
\newblock {GATE}: {A} framework and graphical development environment for
  robust {NLP} tools and applications.
\newblock In \emph{Proc. 40th annual meeting of the association for
  computational linguistics ({ACL} 2002)}, pages 168--175.

\bibitem[{Dickinson et~al.(2012)Dickinson, Brew, and
  Meurers}]{dickinson_language_2012}
Markus Dickinson, Chris Brew, and Detmar Meurers. 2012.
\newblock \emph{Language and computers}.
\newblock John Wiley \& Sons.

\bibitem[{Gunning(2017)}]{gunning_explainable_2017}
David Gunning. 2017.
\newblock Explainable artificial intelligence (xai).
\newblock \emph{Defense Advanced Research Projects Agency (DARPA), nd Web},
  2(2).

\bibitem[{Light et~al.(2005)Light, Arens, and Lu}]{light2005web}
Marc Light, Robert Arens, and Xin Lu. 2005.
\newblock Web-based interfaces for natural language processing tools.
\newblock In \emph{Proceedings of the Second ACL Workshop on Effective Tools
  and Methodologies for Teaching NLP and CL}, pages 28--31.

\end{thebibliography}
\bibliographystyle{acl_natbib}

\appendix

\end{document}